\documentclass{article}
\usepackage{named,amsthm,amsmath,amssymb,proceedings}

\newcommand{\C}[1]{\mathcal{#1}}

\DeclareMathOperator{\NT}{\mathbf{not}}
 \DeclareMathOperator{\ha}{\hookrightarrow}
\DeclareMathOperator{\nha}{\not\!\hookrightarrow}

\newtheorem{thm}{Theorem}[section]

\newtheorem{lem}[thm]{Lemma}

\theoremstyle{definition}
\newtheorem{defn}{Definition}[section]

\theoremstyle{remark}

\newtheorem*{exam}{Example}

\title{Collective Argumentation}
\author{\textbf{Alexander Bochman}\\
Computer Science Department,\\ Holon Academic Institute of Technology, Israel\\ e-mail:
bochmana@hait.ac.il}

\begin{document}

\maketitle

\begin{abstract}
An extension of an abstract argumentation framework, called collective argumentation, is
introduced in which the attack relation is defined directly among sets of arguments. The
extension turns out to be suitable, in particular, for representing semantics of
disjunctive logic programs. Two special kinds of collective argumentation are considered
in which the opponents can share their arguments.
\end{abstract}

\section*{Introduction}

The general argumentation theory \cite{BDKT97,Dun95a} has proved to be a powerful
framework for representing nonmonotonic formalisms in general, and semantics for normal
logic programs, in particular. Thus, it has been shown that the main semantics for the
latter, suggested in the literature, are naturally representable in this framework (see,
e.g., \cite{Dun95,KT99}).

In this report we suggest a certain extension of the general argumentation theory in
which the attack relation is defined directly among sets of arguments. In other words, we
will permit situations in which a set of arguments `collectively' attacks another set of
arguments in a way that is not reducible to attacks among particular arguments from these
sets. It turns out that this extension is suitable for providing semantics for
disjunctive logic programs in which the rules have multiple heads. In addition, it
suggests a natural setting for studying kinds of argumentation in which the opponents
could provisionally share their arguments. Moreover, the original argumentation theory
can be `reconstructed' in this framework by requiring, in addition, that the attack
relation should be \emph{local} in the sense that a set of arguments can attack another
set of arguments only if it attacks a particular argument in this set.

The plan of the paper is as follows. After a brief description of the abstract
argumentation theory, we suggest its generalization in which the attack relation is
defined on sets of arguments. It is shown that the suggested collective argumentation
theory is adequate for representing practically any semantics for disjunctive logic
programs. As an application of the general theory, we consider two special cases of the
general framework in which the opponents can share their arguments. The semantics
obtained in this way will correspond to some familiar proposals, given in the literature.

\section{Abstract Argumentation Theory}

We give first a brief description of the argumentation theory from \cite{Dun95a}.

\begin{defn}
An {\em abstract argumentation theory\/} is a pair $\langle\C{A},\ha\rangle$, where
$\C{A}$ is a set of {\em arguments\/}, while $\ha$ a binary relation of an {\em attack\/}
on $\C{A}$.
\end{defn}

A general task of the argumentation theory consists in determining `good' sets of
arguments that are safe in some sense with respect to the attack relation. To this end,
we should extend first the attack relation to sets of arguments: if $\Gamma$, $\Delta$
are sets of arguments, then $\Gamma\ha\Delta$ is taken to hold iff $\alpha\ha\beta$, for
some $\alpha\in\Gamma$, $\beta\in\Delta$.

An argument $\alpha$ will be called {\em allowable\/} for the set of arguments $\Gamma$,
if $\Gamma$ does not attack $\alpha$. For any set of arguments $\Gamma$, we will denote
by $[\Gamma]$ the set of all arguments allowable by $\Gamma$, that is
$$[\Gamma]\ =\ \{\alpha\mid \Gamma\nha\alpha\}$$

An argument $\alpha$ will be called {\em acceptable\/} for the set of arguments $\Gamma$,
if $\Gamma$ attacks any argument against $\alpha$. As can be easily checked, the set of
arguments that are acceptable for $\Gamma$ coincides with $[[\Gamma]]$.

Using the above notions, we can give a quite simple characterization of the basic objects
of an abstract argumentation theory.
\begin{defn}
A set of arguments $\Gamma$ will be called
\begin{itemize}
\item {\em conflict-free\/} if $\Gamma\subseteq[\Gamma]$;

\item {\em admissible\/} if it is conflict-free and $\Gamma\subseteq[[\Gamma]]$;

\item a {\em complete extension\/} if it is conflict-free and $\Gamma=[[\Gamma]]$;

\item a {\em preferred extension\/} if it is a maximal  complete extension;

\item a {\em stable extension\/} if $\Gamma=[\Gamma]$.
\end{itemize}
\end{defn}

A set of arguments $\Gamma$ is conflict-free if it does not attack itself. A
conflict-free set $\Gamma$ is admissible iff any argument from $\Gamma$ is also
acceptable for $\Gamma$, and it is a complete extension if it coincides with the set of
arguments that are acceptable with respect to it. Finally, a stable extension is a
conflict-free set of arguments that attacks any argument outside it. Clearly, any stable
extension is also a preferred extension, any preferred extension is a complete extension,
and any complete extension is an admissible set. Moreover, as has been shown in
\cite{Dun95a}, any admissible set is included in some complete extension. Consequently,
preferred extensions coincide with maximal admissible sets. In addition, the set of
complete extensions forms a complete lower semi-lattice: for any set of complete
extensions, there exists a unique greatest complete extension that is included in all of
them. In particular, there always exists a least complete extension of an argumentation
theory.

As has been shown in \cite{Dun95}, under a suitable translation, the above objects
correspond to well-known semantics suggested for normal logic programs. Thus, stable
extensions correspond to stable models (answer sets), complete extensions correspond to
partial stable models, preferred extensions correspond to regular models, while the least
complete extension corresponds in this sense to the well-founded semantics (WFS). These
results have shown, in effect, that the abstract argumentation theory successfully
captures the essence of logical reasoning behind normal logic programs.

Unfortunately, the above argumentation theory cannot be extended directly to disjunctive
logic programs. The reasons for this shortcoming, as well as a way of modifying the
argumentation theory are discussed in the next section.

\section{Collective Argumentation}

We begin with pointing out a peculiar discrepancy between the abstract argumentation
theory, on the one hand, and the general abductive framework used for interpreting
semantics for logic programs, on the other hand (see, e.g., \cite{BDKT97,Dun95,KT99}).
The main objects of the abductive argumentation theory are sets of assumptions
(abducibles) of the form $\NT p$ that play the role of arguments in the associated
argumentation theory. In addition, the attack relation is defined in this framework as a
relation between sets of abducibles and particular abducibles they attack. For example,
the program rule $r\leftarrow\NT p,\NT q$ is interpreted as saying that the set of
assumptions $\{\NT p,\NT q\}$ attacks the assumption $\NT r$.

The above attack relation is employed for defining the basic objects (such as extensions)
of the source abductive framework. The abstract argumentation theory defines its main
objects, however, as \emph{sets} of arguments. Consequently, they should correspond to
\emph{sets of sets} of assumptions in the abductive framework. The abductive theory
defines such objects, however, as certain plain sets of assumptions! In other words, we
have a certain discrepancy between the levels of representations of intended objects in
these two theories.

The above discrepancy will disappear once we notice that all the basic objects of the
abstract argumentation theory are definable, in effect, in terms of the derived attack
relation $\Gamma\ha\alpha$ between sets of arguments and particular arguments; only the
latter was used in defining the above operator $[\Gamma]$. As a result, the abductive
argumentation theory can be constructed in the same way as the abstract theory, with the
only distinction that the attack relation between sets of arguments and particular
arguments is not reducible to the attack relation among individual arguments.

The above construction of abductive argumentation naturally suggests that assumptions, or
abducibles, can be considered as full-fledged arguments, while the attack relation is
best describable as a relation among sets of arguments. Indeed, once we allow for a
possibility that a set of arguments can produce a nontrivial attack that is not reducible
to attacks among particular arguments, it is only natural to allow also for a possibility
that a set of arguments could be attacked in such a way that we cannot single out a
particular argument in the attacked set that could be blamed for it. In a quite common
case, for example, we can disprove some conclusion jointly supported by the disputed set
of arguments. The following generalization of the abstract argumentation framework
reflects this idea.

\begin{defn}
A {\em collective argumentation theory\/} is a pair $\langle\C{A},\ha\rangle$, where
$\C{A}$ is a set of arguments, and $\ha$ is an attack relation on finite subsets of
$\C{A}$ satisfying the following {\em monotonicity condition\/}:
\begin{flushleft}
(\textbf{Monotonicity})\qquad If $\Gamma\ha\Delta$, then
$\Gamma\cup\Gamma'\ha\Delta\cup\Delta'$.
\end{flushleft}
\end{defn}

Though the attack relation is defined above only on finite sets of arguments, it can be
naturally extended to arbitrary such sets by imposing the following {\em compactness
property\/}:
\begin{flushleft}
(\textbf{Compactness}) $\ \Gamma\ha\Delta$ only if $\Gamma'\ha\Delta'$, for some finite
$\Gamma'\subseteq\Gamma$, $\Delta'\subseteq\Delta$.
\end{flushleft}

As the reader may notice, the suggested argumentation framework has many properties in
common with ordinary sequent calculus, or consequence relations. Moreover, we will use in
what follows the same agreements for the attack relation as that commonly accepted for
consequence relations. Thus, $\Gamma,\Gamma_1\ha\Delta,\Delta_1$ will have the same
meaning as $\Gamma\cup\Gamma_1\ha\Delta\cup\Delta_1$. Similarly,
$\alpha,\Gamma\ha\Delta,\beta$ will be an alternative notation for
$\{\alpha\}\cup\Gamma\ha\{\beta\}\cup\Delta$, etc.

The argumentation theory from \cite{Dun95} satisfies all the above properties. Moreover,
the above modification of the abstract argumentation theory has already been suggested,
in effect, in \cite{KT99}. However, the attack relation defined in the latter paper
satisfied also a couple of further properties described in the following definition.

\begin{defn}
A collective argumentation theory will be called
\begin{itemize}
\item {\em affirmative\/} if no set of arguments attacks the empty set $\emptyset$;

\item {\em local\/} if it satisfies the following condition:
\begin{flushleft}
(\textbf{Locality})\qquad If $\Gamma\ha\Delta,\Delta'$, then either $\Gamma\ha\Delta$ or
$\Gamma\ha\Delta'$.
\end{flushleft}

\item \emph{normal} if it is both affirmative and local.
\end{itemize}
\end{defn}

If a collective argumentation theory is normal, then it can be easily shown that
$\Gamma\ha\Delta$ will hold if and only if $\Gamma\ha\alpha$, for some $\alpha\in\Delta$.
Consequently, the attack relation in such argumentation theories is reducible to the
relation $\Gamma\ha\alpha$ between sets of arguments and single arguments, and the
resulting argumentation theory will coincide, in effect, with that given already in
\cite{Dun95}.

It turns out, however, that the general, non-local framework of collective argumentation
is precisely what is needed in order to represent semantics of disjunctive logic
programs.

\subsection{Collective Argumentation and Disjunctive Programs}

Despite an obvious success, the abstract argumentation theory is still not abundant with
intuitions and principles that could guide its development independently of applications.
In this respect, logic programming and its semantics constitute one of the crucial
sources and driving forces behind development of argumentation theories. Consequently, as
a first step in studying collective argumentation, we consider its representation
capabilities in describing semantics for disjunctive logic programs.

In what follows, given a set of propositional atoms $C$, we will denote by $\overline C$
the complement of $C$ in the set of all atoms. In addition, $\NT C$ will denote the set
of all negative literals (abducibles) $\NT p$, for $p\in C$.

By the general correspondence between normal logic programs and abductive argumentation
frameworks, a set of abducibles $\NT C$ attacks an abducible $\NT p$ in the abductive
theory associated with a normal logic program $P$ if $P$, taken together with $\NT C$ as
a set of additional assumptions, allows to derive $p$.

The above description immediately suggests a generalization according to which any
disjunctive logic program $P$ determines an attack relation among sets of abducibles as
follows:

\begin{center}
 $\NT C$ attacks $\NT D\ $ iff $\ P\cup\NT C$ derives $\bigvee D$.
\end{center}

As can be easily verified, the above defined attack relation satisfies all the properties
of collective argumentation. However, it is in general not local: $P\cup\NT C$ may
support $p\lor q$ without supporting either $p$ or $q$. Still, it will be affirmative for
disjunctive logic programs without constraints.

The appropriateness of the original argumentation theory for representing semantics of
normal logic programs was based, ultimately, on the fact that these semantics are
completely determined by rules of the form $p\gets\NT C$ that are derivable from a
program. A similar principle, called {\em the principle of partial deduction, or
evaluation\/} is valid also for the majority of semantics suggested for disjunctive logic
programs. According to this principle, semantics of such programs should be completely
determined by rules $C\gets\NT D$ without positive atoms in bodies that are derivable
from the source program. See also \cite{BoJLP} for the role of this principle in
determining semantics of logic programs of a most general kind.

The above considerations indicate that practically all `respectable' semantics for
disjunctive programs should be expressible in terms of collective argumentation theories
associated with such programs.

It turns out, however, that the actual semantics suggested for disjunctive programs do
not fit easily into the general constructions of Dung's argumentation theory. A most
immediate reason for this is that the operator $[\Gamma]$ of the abstract argumentation
theory is no longer suitable for capturing the main content of the collective attack
relation, since the latter is defined as holding between sets of arguments. Accordingly,
it seems reasonable to generalize it to an operator that outputs a set of sets of
arguments:
\[\langle\Gamma\rangle=\{\Delta\mid\Gamma\nha\Delta\}\]

As in the abstract argumentation theory, $\langle\Gamma\rangle$ will collect argument
sets that are \emph{allowable} with respect to $\Gamma$. Notice that, due to monotonicity
of the attack relation, $\langle\Gamma\rangle$ will be closed with respect to subsets,
that is, if $\Delta\in\langle\Gamma\rangle$ and $\Phi\subseteq\Delta$, then
$\Phi\in\langle\Gamma\rangle$. Consequently, any set $\langle\Gamma\rangle$ will be
completely determined by maximal argument sets belonging to it. As can be easily
verified, such maximal sets will always exist due to compactness of the attack relation.

\subsection{Stable and Partial Stable Argument Sets}

Using the above generalized operator of allowability, we can give a rather simple
description of stable and partial stable models for disjunctive programs (see, e.g.,
\cite{GL91,Prz91b}) in terms of collective argumentation.

\begin{defn}
\begin{itemize}
\item A set of arguments $\Gamma$ will be said to be \emph{stable} with respect to a
collective argumentation theory if it is a maximal set in $\langle\Gamma\rangle$.

\item A pair of sets $(\Gamma,\Delta)$ will be called \emph{p-stable} if
$\Gamma\subseteq\Delta$, $\Gamma$ is a maximal set in $\langle\Delta\rangle$, and
$\Delta$ is a maximal set in $\langle\Gamma\rangle$.
\end{itemize}
\end{defn}

The following lemmas give more direct, and often more convenient, descriptions of the
above objects. The proofs are immediate, so we omit them.

\begin{lem}
$\Gamma$ is a stable set iff $\Gamma=\{\alpha\mid \Gamma\nha\Gamma,\alpha\}$.
\end{lem}

The above equation says that a stable set is a set $\Gamma $ consisting of all arguments
$\alpha $ such that $\Gamma $ does not attack $\Gamma\cup\{\alpha \}$. A similar
description can be given for partial stable sets:

\begin{lem}
$(\Gamma,\Delta)$ is p-stable iff $\Gamma\subseteq\Delta$,
$\Delta=\{\alpha\mid\Gamma\nha\Delta,\alpha\}$, and $\Gamma=\{\alpha\mid
\Delta\nha\Gamma,\alpha\}$.
\end{lem}

Recall that normal collective argumentation theories could be identified with abstract
Dung's argumentation theories. Moreover, the above descriptions can be used to show that
if a collective argumentation theory is normal, then stable argument sets will coincide
with stable extensions, while p-stable pairs will correspond exactly to complete
extensions of the abstract argumentation theory. These facts could also be obtained as a
by-product of the correspondence between such objects and relevant semantics of
disjunctive programs stated below.

The correspondence between the above descriptions and (partial) stable models of
disjunctive logic programs is established in the following theorem.

\begin{thm}
If $\C{A}_P$ is a collective argumentation theory corresponding to a disjunctive program
$P$, then
\begin{itemize}
\item $C$ is a stable model of $P$ iff $\NT\overline C$ is a stable set in $\C{A}_P$.
\item $(C,D)$ is a p-stable model of $P$ iff $(\NT\overline C,\NT\overline D)$ is
p-stable in $\C{A}_P$.
\end{itemize}
\end{thm}

P-stable models have been introduced in \cite{BoJLP} as a slight modification of
Przymusinski's partial stable models for disjunctive programs from \cite{Prz91b}; the
reason for the modification was that the original Przymusinski's semantics violated the
above-mentioned principle of partial deduction. In our present context, this means that
it is not definable directly in terms of the collective argumentation theory associated
with a disjunctive program. Note, however, that the modification does not change the
correspondence with partial stable models for normal logic programs.

The above results could serve as an instance of our earlier claim that semantics of
disjunctive programs are representable in the framework of collective argumentation.
These results reveal, however, that the relevant objects are significantly different from
the corresponding objects of the abstract argumentation theory. In order to get a further
insight on the differences, we will consider now various notions of admissibility for
argument sets that are definable in the framework of collective argumentation.

\subsection{Argument sharing}

In ordinary disputation and argumentation the parties can provisionally accept some of
the arguments defended by their adversaries in order to disprove the latter. Two basic
cases of such an `argument sharing' in attacking the opponents are described in the
following definition (see also \cite{BDKT97}).

\begin{defn}
\begin{itemize}
\item $\Gamma$ {\em positively attacks\/} $\Delta$ (notation $\Gamma\ha^+\Delta$) if
$\Gamma,\Delta\ha\Delta$;

\item $\Gamma$ {\em negatively attacks\/} $\Delta$ (notation $\Gamma\ha^-\Delta$) if
$\Gamma\ha\Gamma,\Delta$.
\end{itemize}
\end{defn}

In a positive attack, the proponent temporarily accepts opponent's arguments in order to
disprove the latter, while in a negative attack she shows that her arguments are
sufficient for challenging an addition of opponent's arguments. Clearly, if $\Gamma$
attacks $\Delta$ directly, then it attacks the latter both positively and negatively. The
reverse implications do not hold, however.

Note that the above defined notion of a stable argument set was formulated, in effect, in
terms of negative attacks. Indeed, it is easy to see that a set $\Gamma$ is stable iff it
negatively attacks any argument outside it:
\[\Gamma=\{\alpha\mid \Gamma\nha^-\alpha\}\]

As can be seen, the above definition is equivalent to the definition of stable extensions
in the abstract argumentation theory, given earlier, so stable extensions and stable sets
of collective argumentation are indeed close relatives.

Recall now that admissible argument sets in Dung's argumentation theory are definable as
conflict-free sets that counterattack any argument against them. Given the above
proliferation of the notion of an attack in collective argumentation, however, we can
obtain a number of possible definitions of admissibility by allowing different kinds of
attack and/or counterattack among sets of arguments. Three such notions turns out to be
of special interest.

\begin{defn}
A conflict-free set of arguments $\Gamma$ will be called
\begin{itemize}
\item \emph{admissible} if $\Gamma\ha\Delta$ whenever $\Delta\ha\Gamma$;

\item \emph{positively admissible} if $\Gamma\ha^+\Delta$ whenever $\Delta\ha^+\Gamma$;

\item \emph{negatively admissible} if $\Gamma\ha^-\Delta$ whenever $\Delta\ha^-\Gamma$.
\end{itemize}
\end{defn}

Plain admissibility is a direct counterpart of the corresponding notion from the abstract
argumentation theory. Unfortunately, in the context of collective argumentation it lacks
practically all the properties it has in the latter. Notice, in particular, that stable
sets as defined above need not be admissible in this sense.

As can be seen, positive and negative admissibility coincide with plain admissibility for
respective `extended' attack relations. The latter have some specific features that make
the overall structure simpler and more regular. They will be described in the following
sections.

\section{Negative Argumentation}

The definition below provides a general description of collective argumentation based on
a negative attack. Such argumentation theories will be shown to be especially suitable
for studying stable argument sets.

\begin{defn}
A collective argumentation theory will be called {\em negative\/} if
$\Gamma\ha\Gamma,\Delta$ always implies $\Gamma\ha\Delta$.
\end{defn}

As can be easily verified, any collective argumentation theory will be negative with
respect to the negative attack relation $\ha^-$. Moreover, the latter determines a least
negative `closure' of the source attack relation.

The following result gives an important alternative characterization of negative
argumentation; it establishes a correspondence between negative argumentation and
\emph{shift} operations studied in a number of papers on disjunctive logic programming
\cite{DGM94,Sch95,YYG00}.

\begin{lem}
An argumentation theory is negative iff it satisfies:
\begin{flushleft}
(\textbf{Importation})\qquad If $\Gamma\ha\Delta,\Phi$, then $\Gamma,\Delta\ha\Phi$.
\end{flushleft}
\end{lem}

\begin{proof}
If the argumentation theory is negative and $\Gamma\ha\Delta,\Phi$, then
$\Gamma,\Delta\ha\Gamma,\Delta,\Phi$ by monotonicity, and hence $\Gamma,\Delta\ha\Phi$.
The reverse implication is immediate.
\end{proof}

As an important special case of Importation, we have that if $\Gamma\ha\Delta$, then
$\Gamma,\Delta\ha\emptyset$. Thus, any nontrivial negative argumentation theory is bound
to be non-affirmative. Furthermore, this implies that self-contradictory arguments attack
any argument:
$$\text{If }\Delta\ha\Delta\text{, then }\Delta\ha\Gamma$$

These `classical' properties indicate that negative attack is similar to a rule
$A\vdash\neg B$ holding in a supraclassical consequence relation. Though the latter does
not admit contraposition, we nevertheless have that if $A\vdash\neg(B\land C)$, then
$A,B\vdash\neg C$.

The connection between negative argumentation and stable argument sets is based on the
following facts about general collective argumentation.

\begin{thm}
\begin{enumerate}
\item If $\Gamma$ is negatively admissible, and $\Delta$ is a conflict-free set that
includes $\Gamma$, then $\Delta$ is also negatively admissible.

\item Stable sets coincide with maximal negatively admissible sets.
\end{enumerate}
\end{thm}

\begin{proof}
(1) Assume that $\Gamma$ is negatively admissible, $\Gamma\subseteq\Delta$ and
$\Phi\ha^-\Delta$. Then $\Phi\ha^-\Delta,\Gamma$, and hence $\Phi,\Delta\ha^-\Gamma$ by
Importation. Since $\Gamma$ is negatively admissible, we obtain $\Gamma
\ha^-\Phi,\Delta$, and hence $\Gamma ,\Delta \ha^-\Phi$ by Importation. But the latter
amounts to $\Delta\ha^-\Phi$, which shows that $\Delta$ is also negatively admissible.

(2) It is easy to check that any stable set is negatively admissible. Moreover, any
superset of a stable set will not already be conflict-free. Consequently stable sets will
be maximal negatively admissible sets. In the other direction, if $\Gamma$ is a maximal
negatively admissible set and $\alpha\notin\Gamma$, then $\Gamma\cup\{\alpha\}$ will not
be conflict-free by the previous claim, and hence $\Gamma,\alpha\ha\Gamma,\alpha$.
Consequently $\Gamma,\alpha \ha^-\Gamma$, and therefore $\Gamma \ha^-\Gamma ,\alpha $
(since $\Gamma$ is negatively admissible). But the latter implies $\Gamma \ha^-\alpha$,
which shows that $\Gamma $ is actually a stable set.
\end{proof}

Recall now that negatively admissible sets are precisely admissible sets with respect to
the negative attack $\ha^-$. Moreover, in negative argumentation theories admissible sets
will coincide with negatively admissible ones, while any conflict-free set will already
be positively admissible. Accordingly, all nontrivial kinds of admissibility in such
theories will boil down to (negative) admissibility; furthermore, maximal admissible sets
in such argumentation theories will coincide with stable sets.

It can also be easily verified that any collective argumentation theory has the same
stable sets as its negative closure. So, Importation is an admissible rule for
argumentation systems based on stable sets. As a result, negative argumentation theories
suggest themselves as a natural framework for describing stable sets.

Though the above results demonstrate that negatively admissible sets behave much like
logically consistent sets, there is a crucial difference: the empty set $\emptyset$ is
not, in general, negatively admissible. Moreover, an argumentation theory may be
`negatively inconsistent', that is, it may have no negatively admissible sets at all;
this happens precisely when it has no stable sets.

Unfortunately, the above considerations indicate also that negative argumentation is
inappropriate for studying argument sets beyond stable ones. Recall that one of the main
incentives for introducing partial stable and well-founded models for normal programs was
the desire to avoid taking stance on each and every literal and argument. However,
negativity implies that self-contradictory arguments attack any argument whatsoever, so
any admissible set is forced now to counter-attack any such argument. In particular, if
$\Delta\ha\Phi$, then any admissible set should attack $\Delta\cup\Phi$. This means that
complete extensions (and partial stable models) are no longer a viable alternative for
such argumentation systems. This means as well that Importation (and corresponding shift
operations in logic programming) is an appropriate operation only for describing stable
models\footnote{despite some attempts made in this direction -- see, e.g., \cite{YYG00}.
Actually, the same difficulty plagues attempts to define partial stable semantics for
default logic.}.

\section{Positive Argumentation}

The following definition provides a description of argumentation based on a positive
attack.

\begin{defn}
A collective argumentation theory will be called {\em positive\/} if
$\Gamma,\Delta\ha\Delta$ always implies $\Gamma\ha\Delta$.
\end{defn}

Any collective argumentation theory will be positive with respect to the positive attack
relation $\ha^+$. Moreover, the latter determines a least positive extension of the
source attack relation.

\begin{exam}
Consider an argumentation theory containing only two arguments $\alpha$ and $\beta$ such
that $\alpha\ha\alpha$ and $\alpha\ha\beta$.  As can be seen, this argumentation theory
has no extensions, while the corresponding positive theory has a unique extension
$\{\beta\}$.
\end{exam}

Similarly to negative argumentation, positive argumentation can be characterized by the
`exportation' property described in the lemma below:
\begin{lem}
An argumentation theory is positive iff it satisfies:
\begin{flushleft}
(\textbf{Exportation})\qquad If $\Gamma,\Delta\ha\Phi$, then $\Gamma\ha\Delta,\Phi$.
\end{flushleft}
\end{lem}

The above characterization implies, in particular, that self-conflicting arguments are
attacked by any argument:
$$\text{If }\Delta\ha\Delta\text{, then }\Gamma\ha\Delta$$

So, in positive argumentation we are relieved, in effect, from the obligation to refute
self-contradictory arguments. In particular, no allowable argument will be
self-contradictory.

It is interesting to note that positive and negative argumentation are, in a sense,
incompatible on pain of trivialization. Namely, if we combine positive and negative
argumentation, we obtain a symmetric attack relation:
\begin{lem}
If an argumentation theory is both positive an negative, then $\alpha\ha\beta$ always
implies $\beta\ha\alpha$.
\end{lem}

\begin{proof}
If $\alpha\ha\beta$, then $\alpha ,\beta \ha\alpha ,\beta $ by monotonicity.
Consequently, $\alpha ,\beta \ha\alpha $ by negativity and hence $\beta \ha\alpha $ by
positivity.
\end{proof}

In this case we can consider the attack relation $\alpha\ha\beta$ as expressing plain
{\em incompatibility\/} of the arguments $\alpha$ and $\beta$ in a classical logical
sense. In other words, we could treat arguments as propositions and define
$\alpha\ha\beta$ as $\alpha\vDash\neg\beta$.

\subsection{Local Positive Argumentation}

As a matter of fact, positively admissible sets were introduced for normal programs in
\cite{KM91} (see also \cite{KT99}) under the name \emph{weakly stable sets}; maximal such
sets were termed \emph{stable theories}. Accordingly, a study of such objects will amount
to a study of admissible sets in collective argumentation theories that are positive
closures of normal argumentation theories.

Note first that a positive closure of a local argumentation theory need not, in general,
be local. Still, the following property provides a characterization of positive theories
arising from local argumentation theories.

\begin{defn}
A collective argumentation theory will be called {\em l-positive\/} if it is positive and
satisfies
\begin{flushleft}
(\textbf{Semi-locality})\quad If $\Gamma\ha\Delta,\Phi$, then either
$\Gamma,\Delta\ha\Phi$ or $\Gamma,\Phi\ha\Delta$
\end{flushleft}
\end{defn}

The following basic result shows that l-positive argumentation theories are precisely
positive closures of local theories. Though the proof is not trivial, we omit it due to
the lack of space.

\begin{thm}
An argumentation theory is l-positive iff it is a positive closure of some local
argumentation theory.
\end{thm}

Due to the above result, stable theories from \cite{KM91} are exactly representable as
maximal admissible sets in l-positive argumentation theories. It turns out that many
properties of stable theories can be obtained in this abstract setting; the corresponding
descriptions will be given in an extended version of this report. Still, we should
mention that, since l-positive argumentation theories are not local, the corresponding
structure of admissible sets is more complex than in the local case. For example, the set
of admissible sets no longer forms a lower semi-lattice.

\section{Preliminary Conclusions}

Collective argumentation suggests itself as a natural extension of the abstract
argumentation theory. It allows, in particular, to represent and study semantics for
disjunctive logic programs. Speaking generally, it constitutes an argumentation framework
in which the attack relation has structural properties allowing to represent cooperation
and sharing of arguments among the parties.

The present report is very preliminary, however; it only barely scratches the surface of
the vast number of problems and issues that arise in this setting. One of such issues
consists in extending the approach to other, weaker semantics suggested for disjunctive
programs. A more general task amounts, however, to determining general argumentation
principles underlying collective argumentation. This is a subject of an ongoing research.

\bibliographystyle{named}

%\bibliography{mybib,logscott,biconlp,argum}

\end{document}